\def\year{2017}\relax
\documentclass[letterpaper]{article}

\usepackage{amssymb}
\usepackage[cmex10]{amsmath}
\usepackage{multirow}
\usepackage{booktabs}
\usepackage{graphicx}
\usepackage{subfigure}
\usepackage{epstopdf}
\usepackage{algorithm}
\usepackage{xcolor}
\usepackage{algorithmic}
\usepackage{setspace}
\usepackage[font=scriptsize]{caption}
\usepackage{hyperref}

\usepackage{aaai17}
\usepackage{times}
\usepackage{helvet}
\usepackage{courier}
\newcommand{\squishlist}{
 \begin{list}{$\bullet$}
  { \setlength{\itemsep}{0pt}
     \setlength{\parsep}{3pt}
     \setlength{\topsep}{3pt}
     \setlength{\partopsep}{0pt}
     \setlength{\leftmargin}{1.5em}
     \setlength{\labelwidth}{1em}
     \setlength{\labelsep}{0.5em} } }

\newcounter{Lcount}
\newcommand{\squishlisttwo}{
\begin{list}{\arabic{Lcount}. }
{ \usecounter{Lcount}
\setlength{\itemsep}{0pt}
\setlength{\parsep}{0pt}
\setlength{\topsep}{0pt}
\setlength{\partopsep}{0pt}
\setlength{\leftmargin}{2em}
\setlength{\labelwidth}{1.5em}
\setlength{\labelsep}{0.5em} } }

\newcommand{\squishend}{
\end{list} }

\title{Collective Vertex Classification Using Recursive Neural Network}

\begin{document}
%

\author{Qiongkai Xu\\
	The Australian National University\\
	Data61 CSIRO\\
	{\tt Xu.Qiongkai@data61.csiro.au}\\\And
	Qing Wang\\
	The Australian National University\\
	{\tt qing.wang@anu.edu.au}\\\AND
	Chenchen Xu\\
	The Australian National University\\
	Data61 CSIRO\\
	{\tt Xu.Chenchen@data61.csiro.au}\\\And
	Lizhen Qu\\
	The Australian National University\\
	Data61 CSIRO\\
	{\tt Qu.Lizhen@data61.csiro.au}\\
}

\maketitle
\vspace{2mm}
\begin{abstract}
\begin{quote}
Collective classification of vertices is a task of assigning categories to each vertex in a graph based on both vertex attributes and link structure.
Nevertheless, some existing approaches do not use the features of neighbouring vertices properly, due to the noise introduced by these features. 
In this paper, we propose a graph-based recursive neural network framework for collective vertex classification. In this framework, we generate hidden representations from both attributes of vertices and representations of neighbouring vertices via recursive neural networks. Under this framework, we explore two types of recursive neural units, naive recursive neural unit and long short-term memory unit. 
We have conducted experiments on four real-world network datasets. The experimental results show that our framework with long short-term memory model achieves better results and outperforms several competitive baseline methods.

\end{quote}
\end{abstract}

\section{Introduction}
In everyday life, graphs are ubiquitous, e.g. social networks, sensor networks, and citation networks. Mining useful knowledge from graphs and studying properties of various kinds of graphs have been gaining popularity in recent years.
Many studies formulate their graph problems as predictive tasks such as
vertex classification~\cite{london2014collective}, link prediction~\cite{negative_link_prediction}, and graph classification~\cite{niepert2016learning}. 

In this paper, we focus on vertex classification task which studies the properties of vertices by categorising them. Algorithms for classifying vertices are widely adopted in web page analysis, citation analysis and social network analysis~\cite{london2014collective}. Naive approaches for vertex classification use traditional machine learning techniques to classify a vertex only based on the attributes or features provided by this vertex, e.g. such attributes can be words of web pages or user profiles in a social network.
Another series of approaches are \emph{collective vertex classification}, where instances are classified simultaneously as opposed to independently. 
Based on the observation that information of neighbouring vertices may help classifying current vertex, some approaches incorporate attributes of neighbouring vertices into classification process, which however introduce noise at the same time and result in reduced performance \cite{chakrabarti1998enhanced,myaeng2000practical}.
Other approaches incorporate the labels of its neighbours. For instance, the Iterative Classification Approach (ICA) integrates the label distribution of neighbouring vertices to assist classification~\cite{link_based_classification}
and the Label Propagation approach (LP) fine-tunes predictions of the vertex using the labels of its neighbouring vertices~\cite{label_propagation}.
However, labels of neighbouring vertices are not representative enough for learning sophisticated relationships of vertices, while using their attributes directly would involve noise.
We thus need an approach that is capable of capturing information from neighbouring vertices, while in the mean time reducing noise of attributes. Utilizing representations learned from neural networks instead of neighbouring attributes or labels is one of the possible approaches~\cite{schmidhuber2015deep}. As graphs normally provide rich structural information, we exploit the neural networks with sufficient complexity for capturing such structures.

Recurrent neural networks were developed to utilize sequence structures by processing input in order, in which the representation of the previous node is used to generate the representation of the current node. Recursive neural networks exploit representation learning on tree structures. Both of the approaches achieve success in learning representation of data with implicit structures which indicate the order of processing vertices~\cite{tang2015document,tai2015improved}. Following these work, we explore the possibility of integrating graph structures into recursive neural network. However, graph structures, especially cyclic graphs, do not provide such an processing order. 
In this paper, we propose a graph-based recursive neural network (GRNN), which allows us to transform graph structures to tree structures and use recursive neural networks to learn representations for the vertices to classify. This framework consists of two main components, as illustrated in Figure~\ref{fig:framework}:

\begin{figure*}[t]
	\centering
	\includegraphics[scale=0.30]{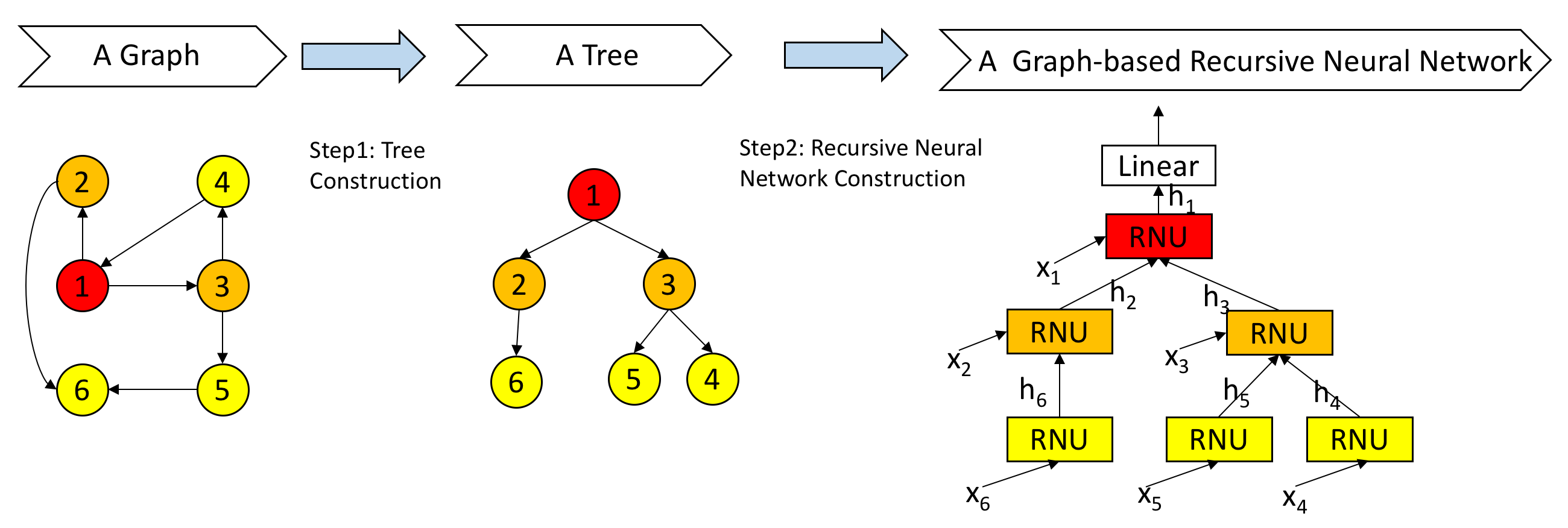}
	\vspace{-1mm}
	\caption{Main components of the Graph-based Recursive Neural Network framework (GRNN): 1) constructing a tree from the vertex to classify (vertex 1); 2) building a recursive neural network on the tree.} \label{fig:framework}
	\vspace{-1mm}
\end{figure*}

\begin{enumerate}
	\item \textbf{Tree construction:} For each vertex to classify $v_t$, we generate a search tree rooted at $v_t$. Starting from $v_t$, we add its neighbouring vertices into the tree layer by layer. 
	\item \textbf{Recursive neural network construction:} We build a recursive neural network for the constructed tree, by augmenting each vertex with one recursive neural unit. The inputs of each vertex are its features and hidden states of its child vertices. The output of a vertex is its hidden states.
\end{enumerate}
Our main contributions are:
(1) We introduce recursive neural networks to solve the collective vertex classification problem. 
(2) We propose a method that can transfer vertices for classification to a locally constructed tree. Based on the tree, recursive neural network can extract representations for target vertices.
(3) Our experimental results show that the proposed approach outperforms several baseline methods. Particularly, we demonstrate that including information from neighbouring vertices can improve performance of classification.


\section{Related Work}

There has been a growing trend to represent data using graphs~\cite{graph_db_survey}. Discovering knowledge from graphs becomes an exciting research area, such as vertex classification~\cite{london2014collective} and graph classification~\cite{niepert2016learning}. Graph classification analyzes the properties of the graph as a whole, while  vertex classification focuses on predicting labels of vertices in the graph. In this paper, we discuss the problem of vertex classification.
The main-stream approaches for vertex classification are
collective vertex classification~\cite{link_based_classification}  which classify vertices using information provided by neighbouring vertices. 
Iterative classification approach~\cite{link_based_classification} models neighbours' label distribution as link features to facilitate classification. Label propagation approach~\cite{label_propagation} assigns a probabilistic label for each vertex and then fine-tunes the probability using graph structure.
However, labels of neighbouring vertices are not representative enough to include all useful information.
Some researchers tried to introduce attributes from neighbouring vertices to improve classification performance. Nevertheless, as reported in~\cite{chakrabarti1998enhanced,myaeng2000practical}, naively incorporating these features may reduce the performance of classification, when original features of neighbouring vertices are too noisy.

Recently, some researchers analysed graphs using deep neural network technologies. Deepwalk~\cite{perozzi2014deepwalk} is an unsupervised learning algorithm to learn vertex embeddings using link structure, while content of each vertex is not considered. Convolutional neural network for graphs~\cite{niepert2016learning} learns feature representations for the graphs as a whole.
Recurrent neural collective classification~\cite{monner2013recurrent} encodes neighbouring vertices via a recurrent neural network, which is hard to capture the information from vertices that are more than several steps away. 

Recursive neural networks (RNN) are a series of models that deal with tree-structured information.  RNN has been implemented in natural scenes parsing~\cite{socher2011parsing} and tree-structured sentence representation learning~\cite{tai2015improved}. Under this framework, representations can be learned from both input features and representations of child nodes. 
Graph structures are more widely used and more complicated than tree or sequence structures. Due to the lack of notable order for processing vertices in a graph, few studies have investigated the vertex classification problem using recursive neural network techniques.
The graph-based recursive neural network framework proposed in this paper can generate the processing order for neural network according to the vertex to classify and the local graph structure.


\section{Graph-based Recursive Neural Networks }
In this section, we present the framework of \emph{Graph-based Recursive Neural Networks} (GRNN). 
A graph $G=(V,E)$ consists of a set of vertices $V = \{v_1, v_2, \dots, v_N\}$ and a set of edges $E\subseteq V\times V$. Graphs may contain cycles, where a cycle is a path from a vertex back to itself.
Let $\mathcal{X} = \{x_1, x_2, \cdots, x_N\}$ be a set of feature vectors, where each $x_i \in \mathcal{X}$ is associated with a vertex $v_i\in V$, $\mathcal{L}$ be a set of labels, and $v_t\in V$ be a vertex to be classified, called \emph{target vertex}. Then, the collective vertex classification problem is to predict the label $y_t$ of $v_t$, such that 
\begin{equation}
	\hat{y_t} = \arg \max_{y_t\in \mathcal{L}} {P_\theta (y_t|v_t, G, \mathcal{X} )}
\end{equation}
using a recursive neural network with parameters $\theta$.

\subsection{Tree Construction}

In a neural network, neurons are arranged in layers and different layers are processed following a predefined order. For example, recurrent neural networks process inputs in sequential order and recursive neural networks deal with tree-structures in a bottom-up manner. However, graphs, particularly cyclic graphs, do not have an explicit order. How to construct an ordered structure from a graph is challenging.

Given a graph $G=(V, E)$, a target vertex $v_t\in V$ and tree depth $d\in \mathbb{N}$, we can construct a tree $T=(V_T, E_T)$ rooted at $v_t$ using breadth-first-search, where $V_T$ is a vertex set, $E_T$ is an edge set, and $(v,w)\in E_T$ means an edge from parent vertex $v$ to child vertex $w$.
The depth of a vertex $v$ in $T$ is the length of the path from $v_t$ to $v$, denoted as $T$.depth($v$). The depth of a tree is maximum depth of vertices in $T$. We use $G$.outgoingVertices($v$) to denote a set of outgoing vertices from $v$, i.e. $\{w|(v,w) \in G\}$.
The tree construction algorithm is described in Algorithm~\ref{alg:tree_gen}. Firstly, a first-in-first-out queue (Q)  is initialized with $v_t$ (lines 1-3). The algorithm iteratively check the vertices in Q. If there is a vertex whose depth is less than $d$, we pop it out from Q (line 6), add all its neighbouring vertices in $G$ as its children in $T$ and push them to the end of Q (lines 9-11). 

\begin{algorithm}[t]
	\caption{Tree Construction Algorithm}
	\label{alg:tree_gen}
	\algsetup{linenodelimiter=.}
	\begin{algorithmic}[1]
		\REQUIRE Graph $G$, Target vertex $v_t$, Tree depth $d$\\
		\ENSURE Tree $T = (V_T, E_T)$
		\STATE Let $Q$ be a queue
		\STATE $V_T = \{v_t\}$, $E_T=\emptyset$
		\STATE $Q$.push($v_t$)
		\WHILE{$Q$ is not empty}
		\STATE $v$ = $Q$.pop()
		\IF{$T$.depth($v$) $\le d$}
		\FOR{ $w$ in $G$.outgoingVertices(v)}
		\STATE // add $w$ to $T$ as child of $v$
		\STATE $V_T = V_T\cup\{w\}$
		\STATE $E_T = E_T\cup\{(v,w)\}$
		\STATE $Q$.push($w$)
		\ENDFOR
		\ENDIF
		\ENDWHILE
		\RETURN $T$
	\end{algorithmic}
\end{algorithm}

\begin{figure}[h]
	\centering
	\includegraphics[scale=0.28]{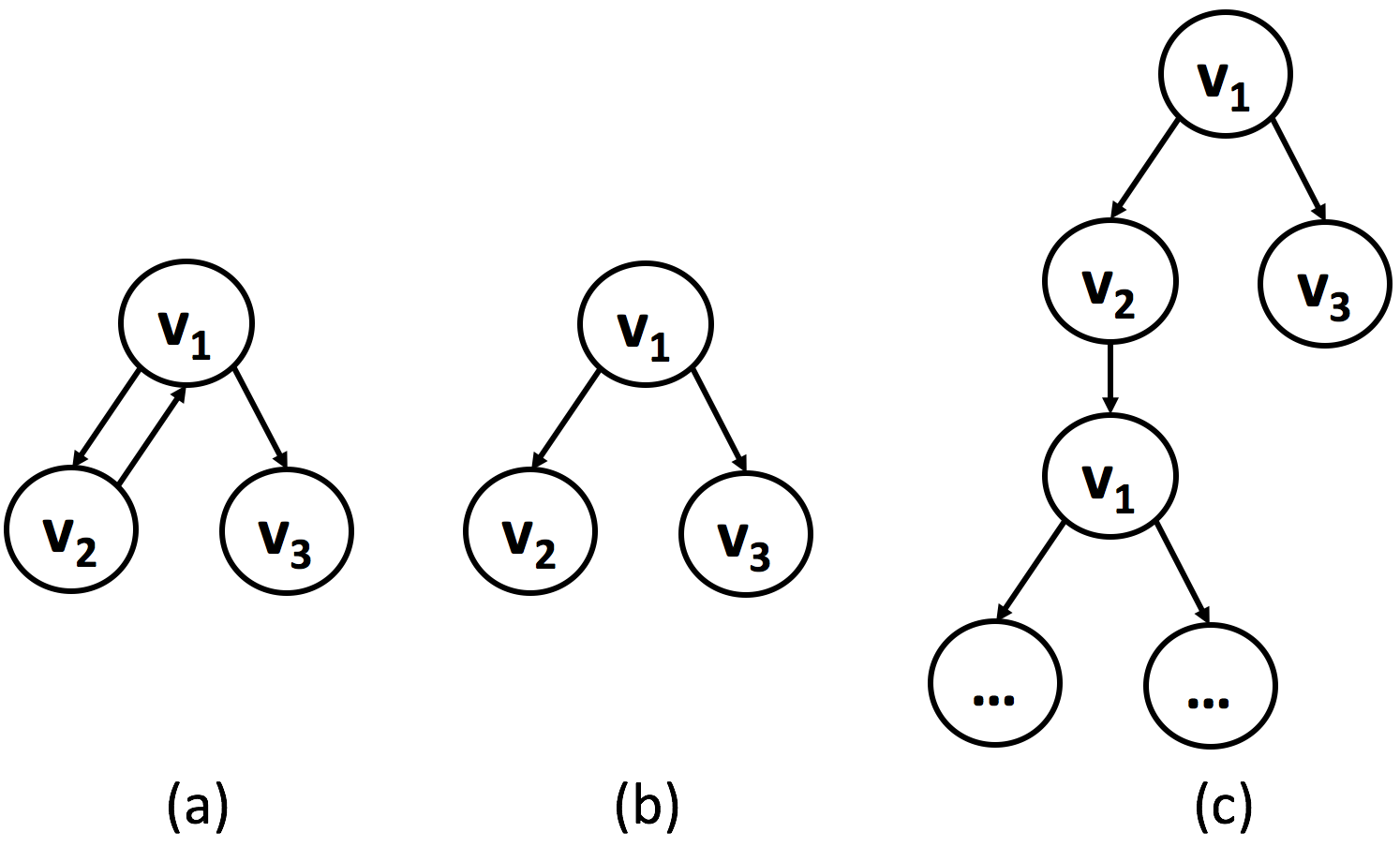}
	\vspace{-1mm}
	\caption{(a) A graph with a cycle; (b) A tree constructed without duplicate vertices; (c) A tree constructed with duplicate vertices.} \label{fig:search_tree}
	\vspace{-1mm}
	
\end{figure}

In general, there are two approaches to deal with cycles in a graph. One is to remove vertices that have been visited, and the other is to keep duplicate vertices.
Fig~\ref{fig:search_tree}.a describes an example of a graph with a cycle between $v_1$ and $v_2$.
Let us start with the target vertex $v_1$. In the first approach, there will be no child vertex for $v_2$, since $v_1$ is already visited. The corresponding tree is shown in Fig~\ref{fig:search_tree}.b. In the second approach, we will add $v_1$ as a child vertex to $v_2$ iteratively and terminate after certain steps. The corresponding tree is illustrated in Fig~\ref{fig:search_tree}.c. When generating the representation of a vertex, say $v_2$, any information from its neighbours may help. We thus include $v_1$ as a child vertex of $v_2$. In this paper, we use the second manner for tree construction.

\subsection{Recursive Neural Network Construction}
Now we construct a recursive neural unit (RNU) for each vertex $v_k\in T$. Each RNU takes a feature vector $x_k$ and hidden states of its child vertices as input. 
We explore two kinds of recursive neural units which are discussed in~\cite{socher2011parsing,tai2015improved}.

\subsubsection{Naive Recursive Neural Unit (NRNU)}\ \\
Each NRNU for a vertex $v_k \in T$ takes a feature vector $x_k$ and the aggregation of the hidden states from all children of $v_k$. The transition equations of NRNU are given as follows:
\begin{equation}
	\widetilde{h_k} = \max_{v_r\in C(v_k)} \{h_r\}
	\vspace{-2mm}
	\label{eq:max_pooling}
\end{equation}
\begin{equation}
	h_k  = \tanh(W^{(h)}x_k + U^{(h)} \widetilde{h_k} + b^{(h)})
\end{equation}

\noindent where  $C(v_k)$ is the set of child vertices of $v_k$, $W^{(h)}$ and $U^{(h)}$ are weight matrix, and $b^{(h)}$ is the bias. The generated hidden state $h_k$ of $v_k$ is related to the input vector $x_k$ and aggregated hidden state $\widetilde{h_k}$.  Different from summing up all hidden states as in~\cite{tai2015improved}, we use max pooling for $\widetilde{h_k}$ (see Eq~\ref{eq:max_pooling}). This is because, in real-life situations, the number of neighbours for a vertex can be very large and some of them are irrelevant for the vertex to classify~\cite{tweet_dataset_neighbour_number}.
We use G-NRNN to refer to the graph-based naive recursive neural network which incorporates NRNU as recursive neural units.

\subsubsection{Long Short-Term Memory Unit (LSTMU)}\ \\
LSTMU is one variation on RNU, which can handle the long term dependency problem by introducing memory cells and gated units~\cite{lstm}.
LSTMU is composed of an input gate $i_k$, a forget gate $f_k$, an output gate $o_k$, a memory cell $c_k$ and a hidden state $h_k$.
The transition equations of LSTMU are given as follows:
\begin{equation}
	\widetilde{h_k} = \max_{v_r\in C(v_k)} \{h_r\}
	\vspace{-1mm}
\end{equation}
\begin{equation}
	i_k = \sigma(W^{(i)}x_k + U^{(i)} \widetilde{h_k} + b^{(i)})
\end{equation}
\begin{equation}
	f_{kr}  = \sigma(W^{(f)}x_k + U^{(f)} h_r + b^{(f)})
\end{equation}
\begin{equation}
	o_k  = \sigma(W^{(o)}x_k + U^{(o)} \widetilde{h_k} + b^{(o)})
\end{equation}
\begin{equation}
	u_k  = \tanh(W^{(u)}x_k + U^{(u)} \widetilde{h_k} + b^{(u)})
\end{equation}
\begin{equation}
	c_k = i_k \odot u_k + \sum_{v_r\in C(v_k)}f_{kr} \odot c_r
	\vspace{-1mm}
\end{equation}

\begin{equation}
	h_k  = o_k \odot \tanh(c_k)
\end{equation}
where  $C(v_k)$ is the set of child vertices of $v_k$, $v_r\in C(v_k)$, $x_k$ is corresponding feature vector of the child vertex $v_k$, $\odot$ is element-wise multiplication and $\sigma$ is sigmoid function, $W^{(*)}$ and $U^{(*)}$ are weight matrices, and $b^{(*)}$ are the biases. $\widetilde{h_k}$ is a vector aggregated from the hidden states of the child vertices.
We use G-LSTM to refer to the graph-based long short-term memory network which incorporates LSTMU as recursive neural units.

After constructing GRNN, we calculate the hidden states of all vertices in $T$ from leaves to root, then we use a softmax classifier to predict label $y_t$ of the target vertex $v_t$ using its hidden states $h_t$ (see Eq~\ref{eq:softmax}).
\begin{equation}
	P_\theta (y_t|v_t, G, \mathcal{X}) = softmax(W^{(s)} h_t + b^{(s)})
	\label{eq:softmax}
\end{equation} 
\begin{equation}
	\hat{y_t} = \arg \max_{y_t\in \mathcal{L}} P_\theta(y_t|v_t, G, \mathcal{X})
\end{equation}
Cross-entropy 
$
J(\theta) = -\frac{1}{N}\sum_{t=1}^{N}\log P_\theta(y_t|v_t, G, \mathcal{X})
$ is used as cost function,
where N is the number of vertices in a training set.

\section{Experimental Setup}
To verify the effectiveness of our approach, we have conducted experiments on four datasets and compared our approach with three baseline methods. We will describe the datasets, baseline methods, and experimental settings.
\subsection{Datasets}
We have tested our approach on four real-world network datasets.
\begin{itemize}
	\item \textbf{Cora}~\cite{cora_dataset} is a citation network dataset which is composed of 2708 scientific publications and 5429 citations between publications. All publications are classified into seven classes:
	\textit{Rule Learning (RU)}, \textit{Genetic Algorithms (GE)}, \textit{Reinforcement Learning (RE)}, \textit{Neural Networks (NE)}, \textit{Probabilistic Methods (PR)}, \textit{Case Based (CA)} and \textit{Theory (TH)}.
	\item \textbf{Citeseer}~\cite{citeseer_dataset} is another citation network dataset which is larger than Cora. Citeseer is composed of 3312 scientific publications and 4723 citations. All publications are classified into six classes:
	\textit{Agents}, \textit{AI}, \textit{DB}, \textit{IR}, \textit{ML} and \textit{HCI}.
	\item \textbf{WebKB}~\cite{webkb_dataset} is a website network collected from four computer science departments in different universities which consists of 877 web pages, 1608 hyper-links between web pages. All websites are classified into five classes: \textit{faculty}, \textit{students}, \textit{project}, \textit{course} and \textit{other}. 
	\item \textbf{WebKB-sim} is a network dataset which is generated from WebKB based on the cosine similarity between each vertex and its top 3 similar vertices according to their feature vectors~\cite{information_retrieval}. We use same feature vectors as the ones in WebKB. This dataset is used to demonstrate the effectiveness of our framework on datasets which may not have explicit relationship represented as edges between vertices, but can be treated as graphs whose edges are based on some metrics such as similarity of vertices.
\end{itemize}
We use abstracts of publications in Cora and Citeseer, and contents of web pages in WebKB to generate features of vertices. For the above datasets, all words are stemmed first, then stop words and words with document frequency less than 10 are discarded. A dictionary is generated by including all these words. We have 1433, 3793, and 1703 for Cora, Citeseer and WebKB, respectively. Each vertex is represented by a bag-of-words vector where each dimension indicates absence or occurrence of a word in the dictionary of the corresponding dataset\footnote{Cora, Citeseer and WebKB can be downloaded from \href{http://linqs.umiacs.umd.edu/projects//projects/lbc/index.html}{LINQS}. We will publish WebKB-sim along with our code.}.

\subsection{Baseline Methods}
We have implemented the following three baseline methods:
\begin{itemize}
	\item \textbf{Logistic Regression (LR)}~\cite{logistic_regression} predicts the label of a vertex using its own attributes through a logistic regression model. 
	\item \textbf{Iterative classification approach (ICA)}~\cite{link_based_classification,london2014collective} utilizes the combination of link structure and vertex features as input of a statistical machine learning model. We use two variants of ICA: \textbf{ICA-binary} uses the occurrence of labels of neighbouring vertices, \textbf{ICA-count} uses the frequency of labels of neighbouring vertices.
	\item \textbf{Label propagation (LP)}~\cite{label_propagation,london2014collective} uses a statistical machine learning to give a label probability for each vertex, then propagates the label probability to all its neighbours. The propagation steps are repeated until all label probabilities converge.
	
\end{itemize}

To make experiments consistent, logistic regression is used for all statistical machine learning components in ICA and LP. We run 5 iterations for each ICA experiment and 20 iterations for each LP experiment\footnote{According to our preliminary experiments, LP converges slower than ICA.}.

\begin{figure*}[t]
	\centering
	\includegraphics[width=18cm,height=8.4cm]{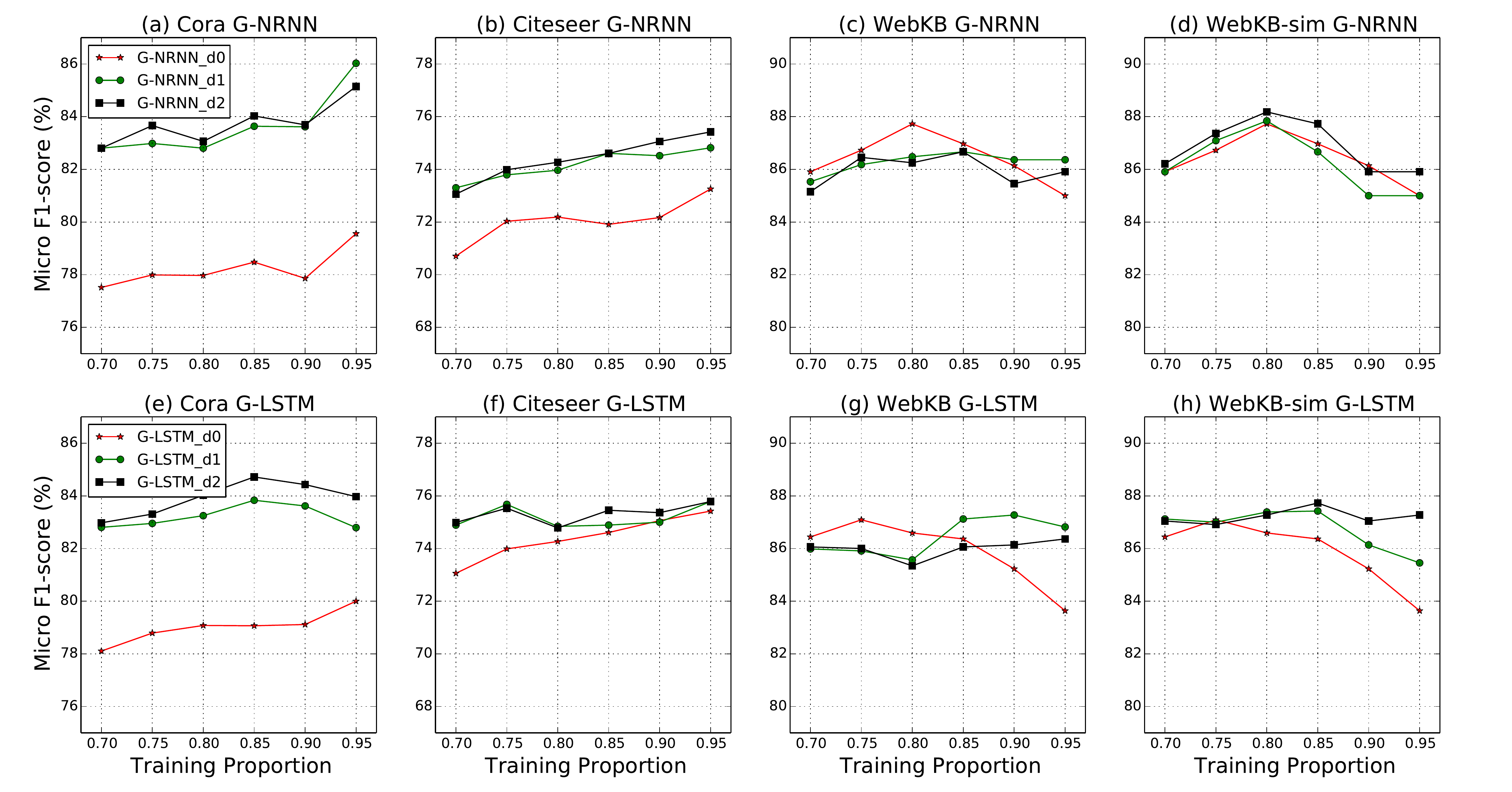}
	\vspace{-1mm}
	\caption{Comparison of G-NRNN and G-LSTM on four datasets: (a) Cora G-NRNN, (b) Citeseer G-NRNN, (c) WebKB G-NRNN, (d) WebKB-sim G-NRNN (e) Cora G-LSTM, (f) Citeseer G-LSTM, (g) WebKB G-LSTM and (h) WebKB-sim G-LSTM.}
	\label{fig:param}
	\vspace{-1mm}
\end{figure*}

\begin{figure*}[t]
	\centering
	\includegraphics[width=18cm,height=4.2cm]{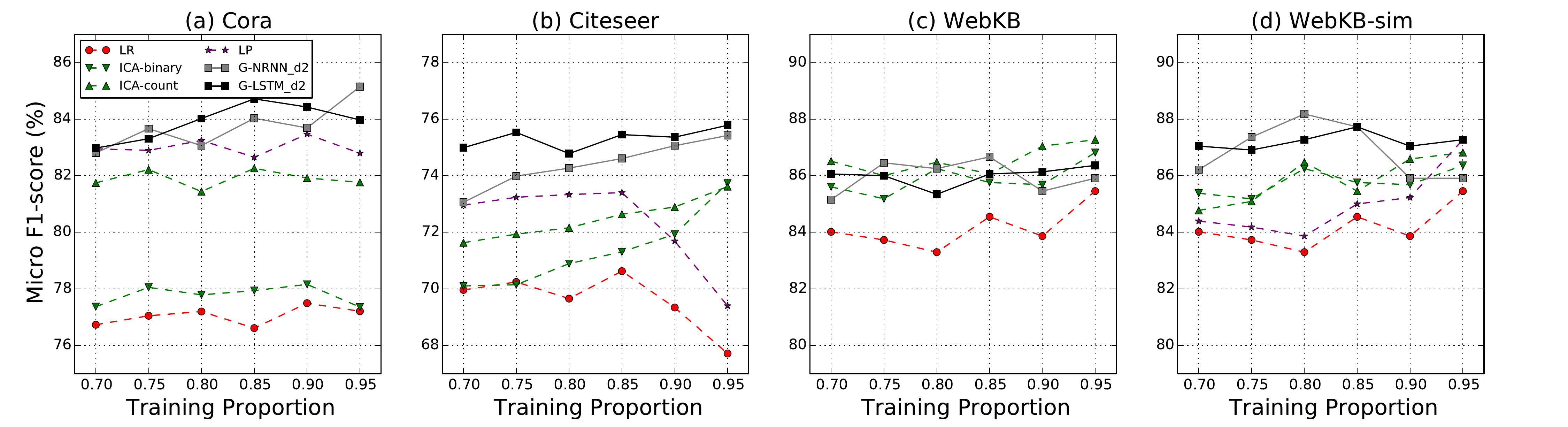}
	\vspace{-1mm}
	\caption{Comparison of G-LSTM with LR, LP and ICA on four datasets: (a) Cora, (b) Citeseer, (c) WebKB and (d) WebKB-sim.}
	\label{fig:baseline_citation}
	\vspace{-1mm}
\end{figure*}

\subsection{Experimental Settings}

In our experiments, we split each dataset into two parts: training set and testing set, with different proportions ($70\%$ to $95\%$ for training). For each proportion setting, we randomly generate 5 pairs of training and testing sets.  For each experiment on a pair of training and testing sets, we run 10 epochs on the training set and record the highest Micro-F1 score~\cite{information_retrieval} on the testing set. Then we report the averaged results from the experiments with the same proportion setting.
According to preliminary experiments, the learning rate is set to 0.1 for LR, ICA and LP and 0.01 for all GRNN models. We empirically set number of hidden states to 200 for all GRNN models. 
Adagrad~\cite{adagrad} is used as the optimization method in our experiments.

\section{Results and Discussion}

\subsection{Model and Parameter Selection}

Figure~\ref{fig:param} illustrates the performance of G-NRNN and G-LSTM on four datasets. We use G-NRNN\_di and G-LSTM\_di to refer to G-NRNN and G-LSTM over trees of depth $i$, respectively, where $i=0,1,2$.

For the experiments on G-NRNN and G-LSTM over trees of different steps, $d1$ and $d2$ outperform $d0$ in most cases\footnote{When $d = 0$, each constructed tree contains only one vertex. }. 
Particularly, the experiments with $d1$ and $d2$ perform better with more than 2\% improvement than $d0$ on Cora, and $d1$ and $d_2$ enjoy a consistent improvement over $d0$ on Citeseer and WebKB-sim. This performance difference is also obvious in WebKB, when the training proportion is larger than 85\%. These mean that introducing neighbouring vertices can improve the performance of classification and more neighbouring information can be obtained by increasing the depth of trees. Using same RNU setting, $d2$ outperforms $d1$ in most experiments on Cora, Citeseer and WebKB-sim. However, for WebKB, $d2$ does not always outperforms $d1$. That is to say, introducing more layers of vertices may help improving the performance, while the choice of the best tree depth depends on applications.

\begin{table}[h!]
	\centering
	\scriptsize
	\caption{Micro-F1 score (\%) of G-LSTM model with different pooling strategies on Cora, Citeseer, WebKB and WebKB-sim.}
	\vspace{-1mm}
	\label{tab:pooling_exp}
	\begin{tabular}{|c|c|c|c|c|c|}
		\hline
		\multirow{ 2}{*}{Method} & Pooling & \multicolumn{4}{|c|}{Datasets}\\ \cline{3-6}
		& Strategy & \ \ \ \ Cora\ \ \ \  \  & \ Citeseer\ \ & \ \ WebKB\ \ \ & WebKB-sim \\
		\hline
		\hline
		& sum 	   & 83.05 & 74.81	&	86.21 & 87.42\\ \cline{2-6}
		G-LSTM\_d1	  & mean	 & \textbf{84.18} & 74.77 & 86.21 & \textbf{87.58}	\\ \cline{2-6}
		& max  		& 83.83& \textbf{74.89}	& \textbf{87.12}& 87.42 \\
		\hline
		& sum 	  & 84.03 &	75.05 & 85.61 &	87.42 \\ \cline{2-6}
		G-LSTM\_d2	  & mean 	& 84.47 &	75.33	&	\textbf{86.21}	&	87.42 \\ \cline{2-6}
		& max 	   & \textbf{84.72} & \textbf{75.45} & 86.06 & \textbf{87.73} \\
		\hline
	\end{tabular}
	\vspace{-1mm}
\end{table}

\begin{figure*}[t]
	\centering
	\includegraphics[scale=0.3]{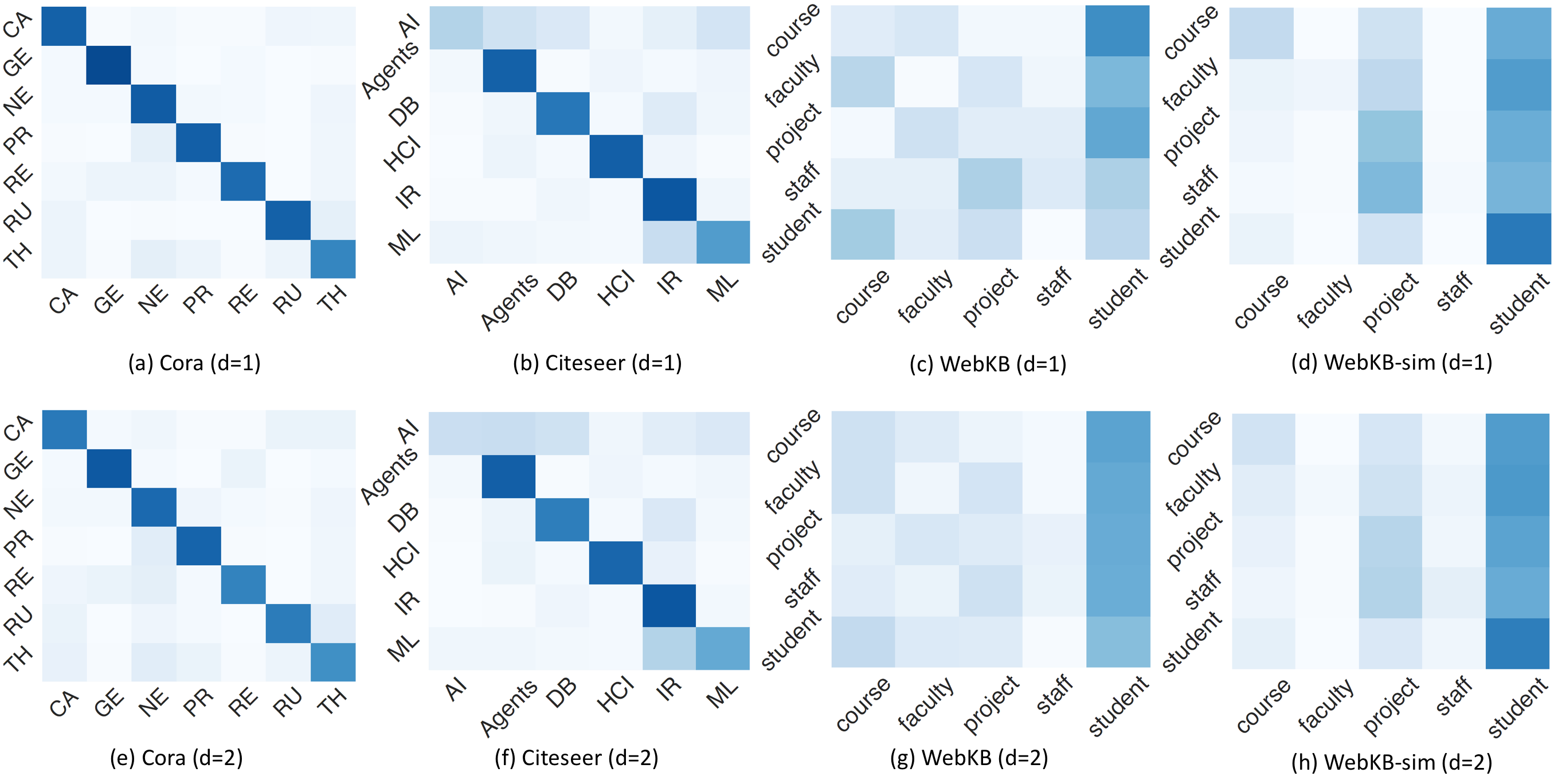}
	\vspace{-1mm}
	\caption{Distribution of label co-occurrence for Cora, Citeseer, WebKB and WebKB-sim, where $d=1, 2$.} \label{heat_map}
	\vspace{-1mm}
\end{figure*}

In Table~\ref{tab:pooling_exp}, we compare three different pooling strategies used in G-LSTM\footnote{As G-NRNN gives similar results, we only illustrate results for G-LSTM here}, \emph{sum}, \emph{mean} and \emph{max} pooling. We use 85\% for training for all datasets here. In general, \emph{mean} and \emph{max} outperform \emph{sum} which is used in~\cite{tai2015improved}. This is probably because, the number of neighbours for a vertex can be very large and summing them up can make $\widetilde{h}$ large for some extreme cases. \emph{max} slightly outperforms \emph{mean} in our experiments, which is probably due to max pooling can select the most influential information of child vertices which filters out noise to some extend.




\subsection{Baseline Comparison}

We compare our approach with the baseline methods on four network datasets. 
As our models with $d_2$ provide better performance, we choose G-NRNN\_d2 and G-LSTM\_d2 as representative models.

In Figure~\ref{fig:baseline_citation}.a and Figure~\ref{fig:baseline_citation}.b, we compare our approach with the baseline methods on two citation networks, Cora and Citeseer. The collective vertex classification approaches, i.e. LP, ICA, G-NRNN and G-LSTM, largely outperform LR which only uses attributes of the target vertex. 
Both G-NRNN\_d2 and G-LSTM\_d2 consistently outperform all baseline methods on the citation networks, which indicates the effectiveness of our approach.
In Figure~\ref{fig:baseline_citation}.c and Figure~\ref{fig:baseline_citation}.d, we compare our approach with the baseline methods on WebKB and WebKB-sim. For WebKB, our method obtains competitive results in comparison with ICA. $LP$ works worse than $LR$ on WebKB, where Micro-F1 score is less than 0.7. For this reason, it is not presented in Figure~\ref{fig:baseline_citation}.c, and we will discuss this in detail in the next subsection. For WebKB-sim, our approaches consistently outperform all baseline methods, and G-LSTM\_d2  outperforms G-NRNN\_d1 when the training proportion is larger than 80\%.

In general, G-LSTM\_d2 outperforms G-NRNN\_d2.
This is likely due to the LSTMU's capability of memorizing information using memory cells. G-LSTM can thus better capture correlations between representations with long dependencies~\cite{lstm}.

\subsection{Dataset Comparison}\label{sec:dataset_comparison}

To analyse co-occurrence of neighbouring labels, we compute the transition probability from target vertices to their neighbouring vertices.
We first calculate the label co-occurrence matrix $M^d$, where $M^d_{i,j}$ indicates co-occurred times of labels $l_i$ of target vertices $v_i$  and labels $l_j$ of $d$-step away vertices $v_j$. Then we obtain a transition probability matrix $T^d$, where
$T^d_{i,j} = M^d_{i,j}/(\sum_{k}M^d_{i,k})$. The heat maps of $T^d$ on four datasets are demonstrated in Figure~\ref{heat_map}.

For Cora and Citeseer, neighbouring vertices tend to share same labels. When $d$ increases to 2, labels are still tightly correlated. That is probably why all ICA, LP G-NRNN and G-LSTM work well on Cora and Citeseer. In this situation, GRNN integrates features of $d$-step away vertices which may directly help classify a target vertex.
For WebKB, correlation of labels is not clear, some label can be strongly related to more than two labels, e.g. \emph{students} connects to \emph{course}, \emph{project} and \emph{student}. Introducing vertices which are more steps away makes the correlation even worse for WebKB, e.g.all labels are most related to \emph{student}. In this situation, LP totally fails, while ICA can learn the correlation of labels that are not same, i.e. \emph{student} may relate to \emph{course} instead of \emph{student} itself. For this dataset, GRNN still achieves competitive results with the best baseline approach. For WebKB-sim, although \emph{student} is still the label with highest frequency, the correlation between labels is clearer than WebKB, i.e. \emph{project} relates to \emph{project} and \emph{student}. That is probably the reason why, the performance of our approach is good on WebKB-sim for both settings and G-LSTM\_d2 achieves better results than G-NRNN\_d2 when the training proportion is larger.

\section{Conclusions and Future work}
In this paper, we have presented a graph-based recursive neural network framework(GRNN) for vertex classification on graphs. We have compared two recursive units, NRNU and LSTMU within this framework. It turns out that LSTMU works better than NRNU on most experiments. Finally, the performance of our proposed methods outperformed several state-of-the-art statistical machine learning based methods.

In the future, we intend to extend this work in several directions. We aim to apply GRNN to large scale graphs. We also aim to improve the efficiency of GRNN and conduct time complexity analysis.

\bigskip

\bibliography{aaai2017}

\begin{thebibliography}{}

\bibitem[\protect\citeauthoryear{Angles and Gutierrez}{2008}]{graph_db_survey}
Angles, R., and Gutierrez, C.
\newblock 2008.
\newblock Survey of graph database models.
\newblock {\em ACM Computing Surveys (CSUR)} 40(1):1.

\bibitem[\protect\citeauthoryear{Baeza-Yates, Ribeiro-Neto, and
  others}{1999}]{information_retrieval}
Baeza-Yates, R.; Ribeiro-Neto, B.; et~al.
\newblock 1999.
\newblock {\em Modern information retrieval}, volume 463.
\newblock ACM press New York.

\bibitem[\protect\citeauthoryear{Chakrabarti, Dom, and
  Indyk}{1998}]{chakrabarti1998enhanced}
Chakrabarti, S.; Dom, B.; and Indyk, P.
\newblock 1998.
\newblock Enhanced hypertext categorization using hyperlinks.
\newblock In {\em ACM SIGMOD Record}, volume~27,  307--318.
\newblock ACM.

\bibitem[\protect\citeauthoryear{Craven \bgroup et al\mbox.\egroup
  }{1998}]{webkb_dataset}
Craven, M.; DiPasquo, D.; Freitag, D.; and McCallum, A.
\newblock 1998.
\newblock Learning to extract symbolic knowledge from the world wide web.
\newblock In {\em Proceedings of the 15th National Conference on Artificial
  Intelligence},  509--516.
\newblock American Association for Artificial Intelligence.

\bibitem[\protect\citeauthoryear{Duchi, Hazan, and Singer}{2011}]{adagrad}
Duchi, J.; Hazan, E.; and Singer, Y.
\newblock 2011.
\newblock Adaptive subgradient methods for online learning and stochastic
  optimization.
\newblock {\em Journal of Machine Learning Research} 12(Jul):2121--2159.

\bibitem[\protect\citeauthoryear{Giles, Bollacker, and
  Lawrence}{1998}]{citeseer_dataset}
Giles, C.~L.; Bollacker, K.~D.; and Lawrence, S.
\newblock 1998.
\newblock Citeseer: An automatic citation indexing system.
\newblock In {\em Proceedings of the 3rd ACM conference on Digital libraries},
  89--98.
\newblock ACM.

\bibitem[\protect\citeauthoryear{Hochreiter and Schmidhuber}{1997}]{lstm}
Hochreiter, S., and Schmidhuber, J.
\newblock 1997.
\newblock Long short-term memory.
\newblock {\em Neural computation} 9(8):1735--1780.

\bibitem[\protect\citeauthoryear{Hosmer~Jr and
  Lemeshow}{2004}]{logistic_regression}
Hosmer~Jr, D.~W., and Lemeshow, S.
\newblock 2004.
\newblock {\em Applied logistic regression}.
\newblock John Wiley \& Sons.

\bibitem[\protect\citeauthoryear{London and
  Getoor}{2014}]{london2014collective}
London, B., and Getoor, L.
\newblock 2014.
\newblock Collective classification of network data.
\newblock {\em Data Classification: Algorithms and Applications} 399.

\bibitem[\protect\citeauthoryear{Lu and
  Getoor}{2003}]{link_based_classification}
Lu, Q., and Getoor, L.
\newblock 2003.
\newblock Link-based classification.
\newblock In {\em Proceedings of the 20th International Conference on Machine
  Learning}, volume~3,  496--503.

\bibitem[\protect\citeauthoryear{McCallum \bgroup et al\mbox.\egroup
  }{2000}]{cora_dataset}
McCallum, A.~K.; Nigam, K.; Rennie, J.; and Seymore, K.
\newblock 2000.
\newblock Automating the construction of internet portals with machine
  learning.
\newblock {\em Information Retrieval} 3(2):127--163.

\bibitem[\protect\citeauthoryear{Monner and Reggia}{2013}]{monner2013recurrent}
Monner, D.~D., and Reggia, J.~A.
\newblock 2013.
\newblock Recurrent neural collective classification.
\newblock {\em IEEE transactions on neural networks and learning systems}
  24(12):1932--1943.

\bibitem[\protect\citeauthoryear{Myaeng and Lee}{2000}]{myaeng2000practical}
Myaeng, S.~H., and Lee, M.-h.
\newblock 2000.
\newblock A practical hypertext categorization method using links and
  incrementally available class information.
\newblock In {\em Proceedings of the 23rd ACM International Conference on
  Research and Development in Information Retrieval}.
\newblock ACM.

\bibitem[\protect\citeauthoryear{Niepert, Ahmed, and
  Kutzkov}{2016}]{niepert2016learning}
Niepert, M.; Ahmed, M.; and Kutzkov, K.
\newblock 2016.
\newblock Learning convolutional neural networks for graphs.
\newblock In {\em Proceedings of the 33rd International Conference on Machine
  Learning}.

\bibitem[\protect\citeauthoryear{Perozzi, Al-Rfou, and
  Skiena}{2014}]{perozzi2014deepwalk}
Perozzi, B.; Al-Rfou, R.; and Skiena, S.
\newblock 2014.
\newblock Deepwalk: Online learning of social representations.
\newblock In {\em Proceedings of the 20th ACM SIGKDD International Conference
  on Knowledge Discovery and Data Mining},  701--710.
\newblock ACM.

\bibitem[\protect\citeauthoryear{Schmidhuber}{2015}]{schmidhuber2015deep}
Schmidhuber, J.
\newblock 2015.
\newblock Deep learning in neural networks: An overview.
\newblock {\em Neural Networks} 61:85--117.

\bibitem[\protect\citeauthoryear{Socher \bgroup et al\mbox.\egroup
  }{2011}]{socher2011parsing}
Socher, R.; Lin, C.~C.; Manning, C.; and Ng, A.~Y.
\newblock 2011.
\newblock Parsing natural scenes and natural language with recursive neural
  networks.
\newblock In {\em Proceedings of the 28th International Conference on Machine
  Learning},  129--136.

\bibitem[\protect\citeauthoryear{Tai, Socher, and
  Manning}{2015}]{tai2015improved}
Tai, K.~S.; Socher, R.; and Manning, C.~D.
\newblock 2015.
\newblock Improved semantic representations from tree-structured long
  short-term memory networks.
\newblock In {\em Proceedings of the 53rd Annual Meeting of the Association for
  Computational Linguistic}.
\newblock ACL.

\bibitem[\protect\citeauthoryear{Tang \bgroup et al\mbox.\egroup
  }{2015}]{negative_link_prediction}
Tang, J.; Chang, S.; Aggarwal, C.; and Liu, H.
\newblock 2015.
\newblock Negative link prediction in social media.
\newblock In {\em Proceedings of the Eighth ACM International Conference on Web
  Search and Data Mining},  87--96.
\newblock ACM.

\bibitem[\protect\citeauthoryear{Tang, Qin, and Liu}{2015}]{tang2015document}
Tang, D.; Qin, B.; and Liu, T.
\newblock 2015.
\newblock Document modeling with gated recurrent neural network for sentiment
  classification.
\newblock In {\em Proceedings of the 2015 Conference on Empirical Methods in
  Natural Language Processing},  1422--1432.

\bibitem[\protect\citeauthoryear{Wang and Zhang}{2008}]{label_propagation}
Wang, F., and Zhang, C.
\newblock 2008.
\newblock Label propagation through linear neighborhoods.
\newblock {\em IEEE Transactions on Knowledge and Data Engineering}
  20(1):55--67.

\bibitem[\protect\citeauthoryear{Zhang \bgroup et al\mbox.\egroup
  }{2013}]{tweet_dataset_neighbour_number}
Zhang, J.; Liu, B.; Tang, J.; Chen, T.; and Li, J.
\newblock 2013.
\newblock Social influence locality for modeling retweeting behaviors.
\newblock In {\em Proceeding of the 23rd International Joint Conference on
  Artificial Intelligence}, volume~13,  2761--2767.
\newblock Citeseer.

\end{thebibliography}
\bibliographystyle{aaai}

\end{document}